\documentclass{article}
\usepackage{ijcai25}

\usepackage{times}
\usepackage{soul}
\usepackage{url}
\usepackage[hidelinks]{hyperref}
\usepackage[utf8]{inputenc}
\usepackage[small]{caption}
\usepackage{graphicx}
\usepackage{amsmath}
\usepackage{amsthm}
\usepackage{booktabs}
\usepackage{algorithm}
\usepackage{algorithmic}
\usepackage[switch]{lineno}
\usepackage{natbib}

\usepackage{amsmath}

\usepackage[nowatermark]{fixmetodonotes}

\DeclareMathOperator*{\argmin}{arg\,min}

\title{Causal Graph Recovery in Neuroimaging through Answer Set Programming}

\author{
Mohammadsajad Abavisani$^{1,3}$
\and
Kseniya Solovyeva$^2$\and
David Danks$^4$\and
Vince Calhoun$^{1,2,3}$\And
Sergey Plis$^{2,3}$\\
\affiliations
$^1$Georgia Institute of Technology\\
$^2$Georgia State University\\
$^3$Center for Translational Research in Neuroimaging and Data Science (TReNDS)\\
$^4$University of California San Diego\\
\emails
\{s.abavisani, vcalhoun\}@gatech.edu,
splis@gsu.edu
}

\begin{document}

\maketitle

\begin{abstract}
    Learning graphical causal structures from time series data presents significant challenges, especially when the measurement frequency does not match the causal timescale of the system. This often leads to a set of equally possible underlying causal graphs due to information loss from sub-sampling (i.e., not observing all possible states of the system throughout time). Our research addresses this challenge by incorporating the effects of sub-sampling in the derivation of causal graphs, resulting in more accurate and intuitive outcomes. We use a constraint optimization approach, specifically answer set programming (ASP), to find the optimal set of answers. ASP not only identifies the most probable underlying graph, but also provides an equivalence class of possible graphs for expert selection. In addition, using ASP allows us to leverage graph theory to further prune the set of possible solutions, yielding a smaller, more accurate answer set  significantly faster than traditional approaches. We validate our approach on both simulated data and empirical structural brain connectivity, and demonstrate its superiority over established methods in these experiments. We further show how our method can be used as a meta-approach on top of established methods to obtain, on average, $12\%$ improvement in F1 score. In addition, we achieved state of the art results in terms of precision and recall of reconstructing causal graph from sub-sampled time series data. Finally, our method shows robustness to varying degrees of sub-sampling on realistic simulations, whereas other methods perform worse for higher rates of sub-sampling. 
\end{abstract}

\section{Introduction}

Causal inference from functional Magnetic Resonance Imaging (fMRI) data has emerged as a critical endeavor to understand the neural mechanisms underlying cognitive processes and behaviors. Researchers not only seek to identify active brain regions during tasks, but also unravel the causal relationships between these regions, often referred to as "effective connectivity" \citep{friston1994functional}. Graphical causal models, such as causal Bayesian networks, have become a popular framework for this purpose, combining directed graphs with joint probability distributions to model the dependencies between different brain regions \citep{pearl2009causality}. These models adhere to the Causal Markov Condition, which asserts that each node in a causal graph is conditionally independent of its non-descendants given its parents \citep{spirtes2001causation}.

However, applying these models to fMRI data is fraught with challenges, particularly due to the mismatch between the temporal resolution of fMRI and the rapid timescale of neural processes. The typical sampling intervals in fMRI, ranging from one to three seconds, are much slower than the millisecond-level interactions between neurons, leading to significant undersampling \citep{valdes2011effective}. This undersampling often results in the identification of multiple causal graphs that are statistically indistinguishable, forming what is known as a Markov Equivalence Class \citep{pearl2009causality, spirtes2001causation}. These problems are exacerbated by the indirect nature of the BOLD signal, which reflects neural activity through complex and variable hemodynamic responses \citep{handwerker2004variation}.

In addition, the inherent variability in the hemodynamic response across different brain regions and subjects, adds another layer of complexity. Variations in the time-to-peak of the BOLD response can lead to incorrect inferences about the direction of causality, particularly when using methods like Granger causality, which assumes a fixed temporal relationship between cause and effect \citep{david2008identifying, seth2013granger}. Although some studies suggest that Granger causality may be robust to certain variations in the hemodynamic response \citep{seth2013granger}, the combination of measurement noise, undersampling, and hemodynamic variability often undermines the reliability of causal inferences from fMRI data.
\begin{figure}[t]
     \includegraphics [width =1\columnwidth]{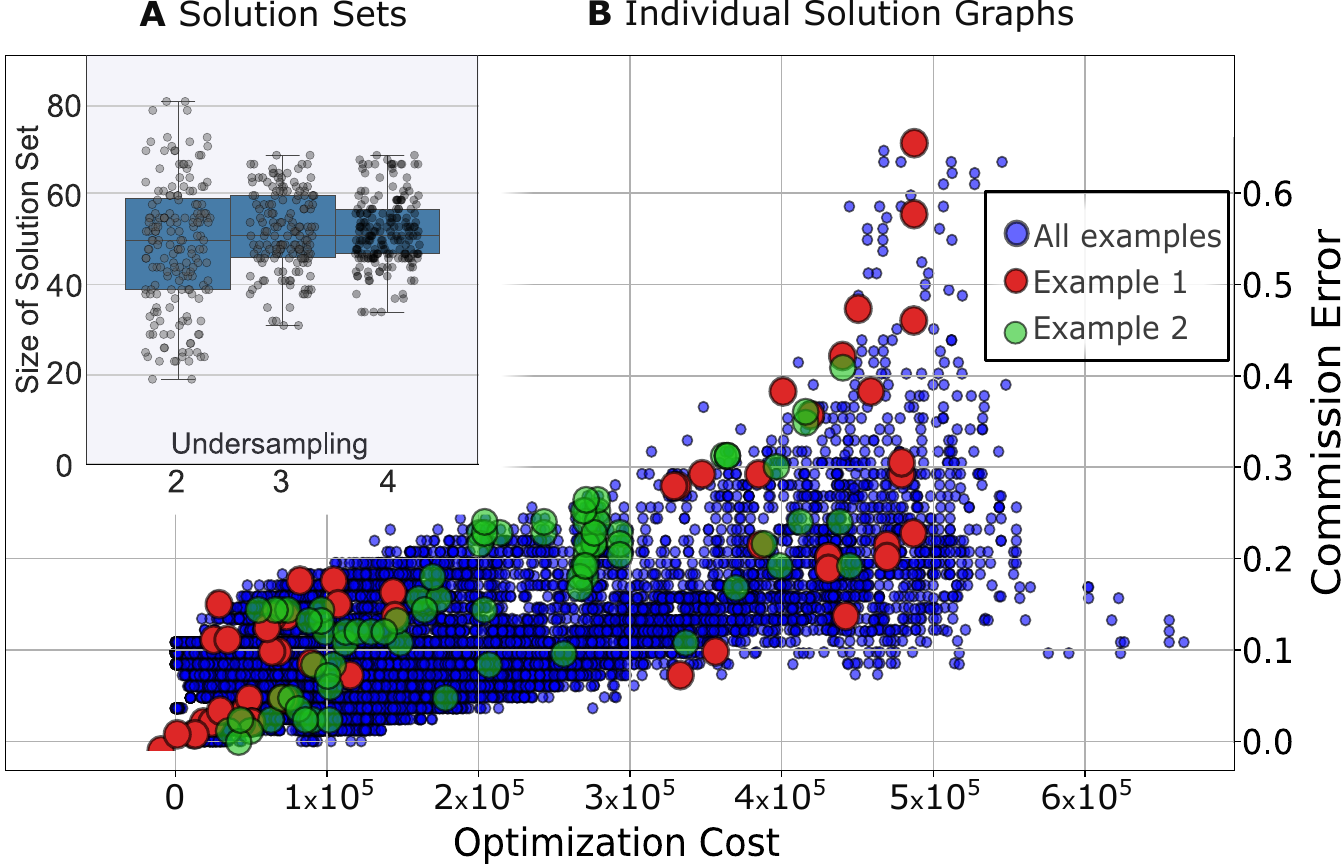}
     \caption{Top left(A): Size of optimization solution set across different undersamplings, repeated 100 times. Bottom right(B): Commission error of the solution vs. optimization cost for that solution. Solutions in one equivalent class are highlighted in red.}
     \label{fig_eq}
\end{figure}

In response to these challenges, this paper proposes a novel approach that explicitly accounts for the effects of undersampling in the derivation of causal graphs. By employing constraint optimization through state-of-the-art Answer Set Programming (ASP), we aim to identify the most probable causal graph from a set of potential candidates. ASP allows for the incorporation of domain-specific knowledge and constraints, facilitating the identification of not only a single graph but an equivalence class of possible graphs, thereby offering a more comprehensive understanding of the underlying causal structure \citep{gebser2012conflict}.

Furthermore, we introduce a method to enhance the robustness and accuracy of causal inference by integrating data from multiple instances of the same underlying causal system, such as combining fMRI data from several subjects performing the same task. This approach synthesizes information across these instances into a singular causal graph, leveraging ASP's ability to handle complex combinatorial problems that would otherwise be infeasible with conventional techniques \citep{lifschitz2008knowledge}.

We validate our approach using both simulated data and real fMRI data from macaque brains, demonstrating its superiority over existing methods \citep{smith2011network, ramsey2010six}. Our results suggest that ASP offers a powerful new tool for causal inference in neuroimaging, providing more accurate and intuitive insights into the brain's functional architecture.

\begin{figure*}[ht]
     \includegraphics [width = 2\columnwidth]{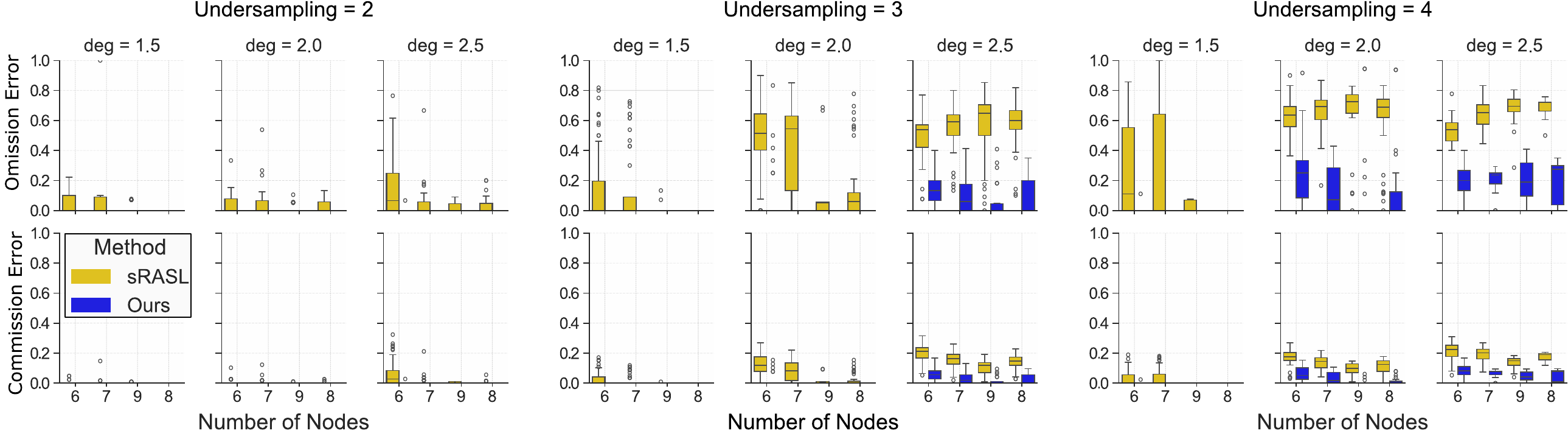}
     \caption{Normalized omission (top) and commission (bottom) errors for edge-breaking experiments with varying undersampling rates and graph densities, comparing the original approach with our improved sRASL-based method.}
     \label{fig_edge_break}
\end{figure*}

\section{Background}
\label{sec:background}
Causal inference from time series data, particularly from fMRI, has been a central focus in computational neuroscience due to its potential to uncover complex neural mechanisms. 
A directed dynamic causal model extends standard causal models \citep{pearl2000models, Spirtes1993} by incorporating temporal dependencies, where the graph $\textbf{G}$ includes nodes representing random variables 
$\textbf{V} = \{V_{1}, V_{2}, \dots, V_{n}\}$ across both the current timestep $t$ ($\textbf{V}^{t}$) and previous timesteps ($\textbf{V}^{t-k}$) that are direct causes of some $V_i^t$. 
These models assume a first-order Markov structure, where $\textbf{V}^{t}$ is conditionally independent of $\textbf{V}^{t-k}$ given $\textbf{V}^{t-1}$ for all $k > 1$, allowing flexibility in the causal timescale \citep{Spirtes2000causation}.

Undersampling presents a significant challenge in fMRI, where data is collected at intervals longer than the timescale of underlying neural processes. If the true causal structure evolves over timepoints 
$\{t^{0}, t^{1}, t^{2}, \dots, t^{k}, \dots\}$, but measurements are taken at $\{t^{0}, t^{u}, t^{2u}, \dots, t^{ku}, \dots\}$, this results in an undersampled graph $\textbf{G}^{u}$ that may obscure the true causal graph 
$\textbf{G}^{1}$ \citep{danks2013learning, gong2015discovering}.

To address the structural implications of undersampling, several approaches have been developed. For instance, \citet{danks2013learning} explored the problem structurally, while \citet{gong2015discovering} provided a parametric approach for two-variable systems. The resulting undersampled graphs are typically compressed, with temporal information encoded in the edges. Specifically: (1) In the undersampled graph $\mathcal{G}^{u}$, an edge $V_i \rightarrow V_j$ exists if and only if there is a directed path of length $u$ from $V_i$ to $V_j$ in the fully observed graph $\mathcal{G}^{1}$, or equivalently, if there exists a directed path from $V_i^{t-u}$ to $V_j^{t}$ in \textbf{G$^{1}$}. This property captures how dependencies persist across multiple timesteps despite undersampling. (2) Similarly, a bidirected edge $V_i \leftrightarrow V_j$ in $\mathcal{G}^{u}$ arises when there exists a common ancestor $V_k$ in $\mathcal{G}^{1}$ that has directed paths to both $V_i$ and $V_j$, with the intermediate connections spanning fewer than $u$ timesteps. This implies that $V_k$ serves as an unobserved confounder, introducing dependencies that persist through undersampling.

Dealing with latent confounders, such as those introduced by undersampling, has led to the development of various graph representations, including Partially-Observed Ancestral Graphs (PAGs) \citep{zhang2008causal} 
and Maximal Ancestral Graphs (MAGs) \citep{richardson2002ancestral}. However, these frameworks often struggle with the complexities introduced by undersampling \citep{mooij2020constraint}. 
Compressed graphs have proven more effective in addressing the challenge of inferring the true causal graph $\textbf{G}^{1}$ from an observed graph $\mathcal{H}$ under unknown undersampling rates.

The Rate-Agnostic Structure Learning (RASL) algorithm \citep{plis2015rate} was introduced to address the challenge of causal inference from undersampled data, placing a strong emphasis on directly tackling undersampling while adopting a rate-agnostic approach that avoids assumptions about the undersampling rate. By systematically considering all possible rates and incorporating stopping rules to eliminate redundant graph exploration, RASL marked a significant advancement in computational efficiency. However, it faced practical limitations in scaling to larger graphs, leaving room for further improvement.

Building on these foundations, recent advancements in causal structure learning have led to the development of generalized rate-agnostic approaches that significantly enhance the scalability and efficiency of these methods. The reformulation of RASL into a constraint satisfaction framework, expressed using a declarative language, has shown considerable promise by enabling more efficient analysis of large graphs called Solver-based RASL (sRASL) \citep{abavisani2023grace}. By incorporating additional constraints based on strongly connected components (SCC) structures, this approach provides a scalable and accurate solution for causal inference from undersampled data, capable of analyzing datasets with over 100 nodes—a substantial improvement over earlier methods that struggled with smaller graphs. The proposed sRASL method achieves a 1000-fold improvement in solving time, while maintaining the same theoretical guarantees as its predecessors. Nonetheless, sRASL exhibits limitations when applied to real-world datasets, particularly in the presence of environmental noise, as its performance gains observed in simulations do not fully translate to noisy, real-world conditions.

In a comprehensive study by \citet{sanchez2019estimating}, various statistical methods were assessed for their effectiveness in recovering the causal structure of systems with feedback from synthetic BOLD time series. 
The study compared established methods, such as Granger causal regression and multivariate autoregressive models, with newer approaches like Fast Adjacency Skewness (FASK) and Two-Step, both of which leverage non-Gaussian features of the BOLD signal. 
Their analysis covered feedback structures, excitatory and inhibitory feedback, and models using macaque structural connectivity as well as human resting-state and task data. Several methods demonstrated over 80\% orientation precision and recall, 
including those involving 2-cycles. However, this study did not account for the fact that BOLD data is inherently undersampled. Therefore, in the current article, we investigated how the algorithms studied in the Sanchez-Romero paper track the structure of graphs with a sparse sample and compared these algorithms with an algorithm based on Answer Set Programming (ASP). 
The FASK algorithm showed good results along with the ASP algorithm.

In recent years, causal discovery in fMRI has advanced significantly, with new methods pushing the boundaries of connectivity estimation by incorporating non-linearities, handling latent confounders, 
and applying complex neural network models. Methods such as LiNGAM \citep{shimizu2006lingam} are well-suited for identifying directed causal relationships in linear, non-Gaussian acyclic models, 
making them highly effective for discovering directed interactions within complex neural networks. Other innovative approaches like EC-GAN \citep{kim2021ecgan} use generative adversarial frameworks 
to infer effective connectivity, while models such as the Amortization Transformer \citep{paul2022amortization} and reinforcement learning algorithms like ActorCritic \citep{mnih2016asynchronous} adapt dynamically 
to the intricate patterns of whole-brain causal structures, as seen in CaLLTiF \citep{salehi2021calltif} and DYNOTEARS \citep{pamfil2020dynotears}.

However, for our study, we selected the methods from Sanchez-Romero’s work—GIMME \citep{gates2010gimme}, MVGC \citep{barnett2009mvgc}, MVAR \citep{bressler2003mvar}, and FASK \citep{sanchez2019fask}—as the primary benchmarks for comparison. 
These methods provide a robust, established foundation for causal inference in fMRI, especially in handling real-world challenges such as feedback loops, individual variability, and multivariate time-series dependencies. 
GIMME is particularly effective in capturing both individual and group-level connectivity, essential for population studies, while MVGC and MVAR offer reliable multivariate approaches for assessing directional dependencies in time series data. 
FASK, on the other hand, improves causal orientation accuracy by leveraging non-Gaussian characteristics, which are beneficial in the context of noisy fMRI data. By comparing our method with these well-established approaches, 
we aim to benchmark against standard, validated techniques, allowing us to clearly assess the relative strengths and limitations of our approach in a field-relevant framework.

Building on these findings, we evaluated the performance of these algorithms when applied to undersampled data, benchmarking them against Answer Set Programming (ASP). Our results offer new insights into the robustness and accuracy of these methods 
in the context of undersampling, a frequent challenge in fMRI analysis.

\section{Methods}

In this study, we present an enhanced version of the sRASL framework~\citep{abavisani2023grace} to improve its applicability to functional brain connectivity analysis using fMRI data. We call this new method Real-world noisy RASL or \textbf{RnR} for short. The original sRASL was developed for structural graph analysis, but was limited by its assumption of a single optimal solution and lacked realism in recovered graph densities. To address these limitations, we propose the following key innovations:
\begin{figure}[t]
     \includegraphics[width=1
\columnwidth]{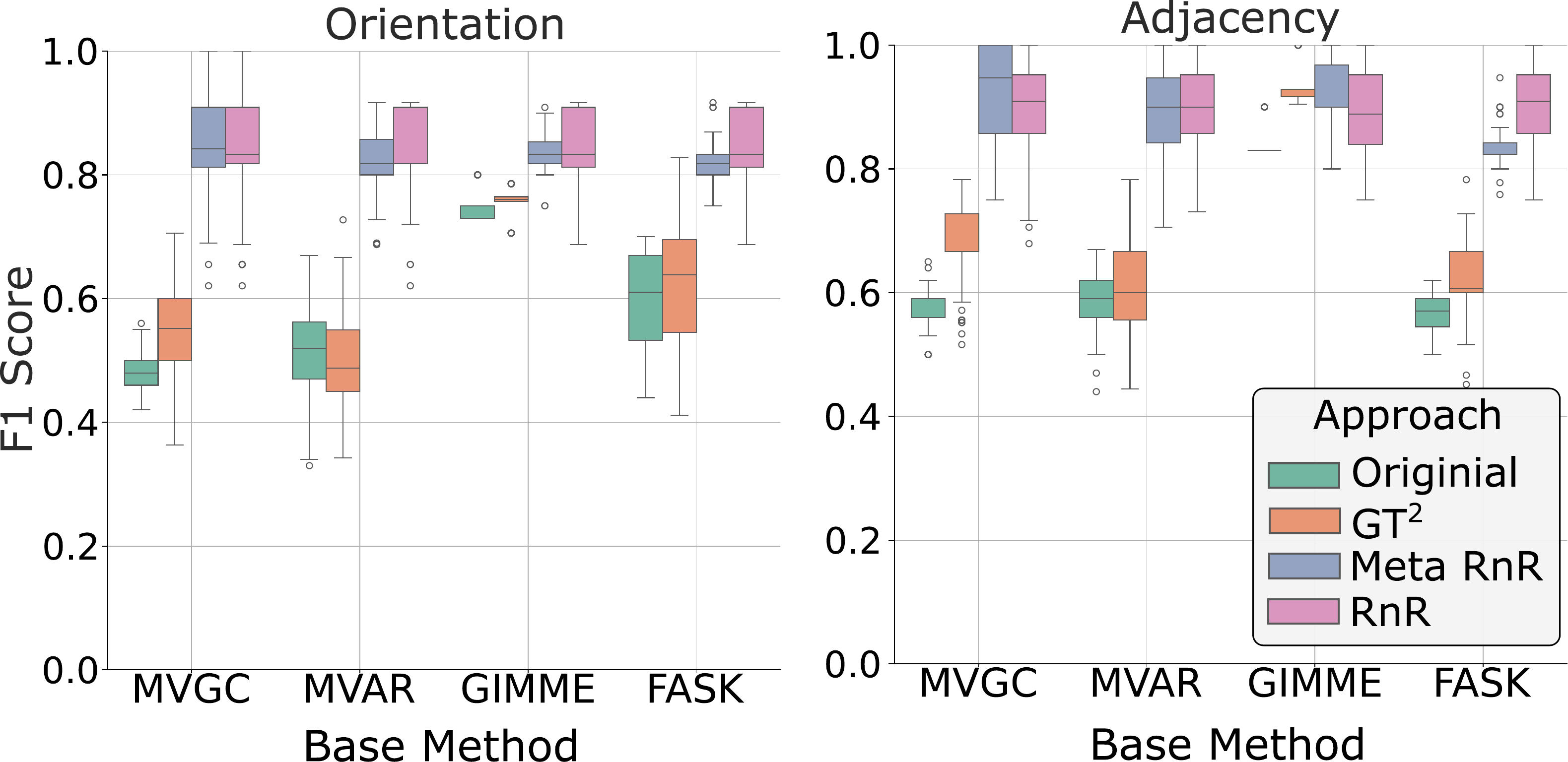}
     \caption{Performance comparison using Sanchez-Romero's data with and without the RASL meta-solver. Applying RASL on top of traditional methods improves accuracy by accounting for undersampling effects, with additional gains achieved by using PCMCI.}
     \label{fig_ruben_all}
\end{figure}

\begin{itemize}
    \item Optimization in OptN Mode: Using \texttt{Clingo} in OptN mode to retrieve all near-optimal solutions, enabling comprehensive exploration of equally valid solutions.
    \item Density Constraints: Introducing a density-matching constraint to ensure that the recovered graph reflects realistic functional brain connectivity.
    \item RnR as Meta Solver: We propose a general framework for using our method as a meta-solver (constraining other causal learning algorithms) to take into account effects of undersampling.
    \item Prioritized Optimization: Implementing a multi-stage optimization process that prioritizes graph density, followed by bidirectional and directed connections.
    \item Adaptive Weighting: Applying adaptive weighting schemes to improve graph accuracy by accounting for uncertainty in fMRI-derived connections.
\end{itemize}

These enhancements collectively enable the recovery of functional connectivity graphs that are both biologically realistic and also structurally accurate, while addressing challenges distinctive to fMRI data.

\subsection{Overview of the sRASL Framework}
The sRASL framework takes as input an undersampled graph $\mathcal{H}$, which can be derived from empirical data $\mathbf{D}$, expert knowledge, or a combination of both. Its key principle relies on the modular structure of strongly connected components (SCCs), which must form a Directed Acyclic Graph (DAG). This structural constraint reduces the solution space by eliminating invalid graph configurations. In its original implementation, sRASL returned only a single optimal solution, limiting its ability to account for inherent noise in fMRI data. Additionally, the absence of density constraints led to unrealistic output graphs (in terms of connectivity densities).

\subsection{Optimization in OptN Mode}
In the original sRASL method, a single solution corresponding to the minimum cost function value was returned. This is restrictive because multiple solutions can have comparable costs, especially in noisy datasets like fMRI. To overcome this, we utilize \texttt{Clingo} in OptN mode, which retrieves all near-optimal solutions within a specified cost range. The optimization objective is defined as:

\begin{equation} \label{optimeq}
\begin{aligned}
\mathcal{G}^* \in \argmin \Bigg(
& \sum_{e \in \mathcal{H}} I[e \notin \mathcal{G}] \cdot w(e \in \mathcal{H}) \\
& + \sum_{e \notin \mathcal{H}} I[e \in \mathcal{G}] \cdot w(e \notin \mathcal{H})
\Bigg)
\end{aligned}
\end{equation}

\noindent where the indicator function $I(c)=1$ if condition $c$ holds, and $0$ otherwise. The weights $w(e \in \mathcal{H})$ and $w(e \notin \mathcal{H})$ reflect the importance of edge presence or absence, respectively. By setting the OptN flag in \texttt{Clingo}, we retrieve all solutions within a practical cost range, enabling a more robust exploration of near-optimal graphs.

\subsection{RnR as meta solver}

A common challenge in causal learning is unobserved common causes. In the present context, if the common cause of two nodes is removed due to lower temporal resolution, then there will be bidirectional correlations between nodes, as described in Section \ref{sec:background}. Many causal structure learning methods produce bidirected edges~\citep{pcmci, granger1969investigating, lutkepohl2005new}, but methods that do not account for undersampling will not produce bidirected edges~\citep{gates2010gimme, barnett2009mvgc, bressler2003mvar, sanchez2019fask}. We present an approach where our method can be used as a meta-solver, thereby gaining the benefits of the chosen first-order method, while accommodating undersampling due to our method. 

In addition, our empirical studies revealed that, when methods return a length-2 loop between two nodes, then there is always (i) a bidirected edge between them in the correct model, and (ii) one or both of the two directed edges. We thus add bidirected edges to the graph output by the first-order method. We also add both directed edges to the graph, though with low weight for each direction to encode uncertainty about which might be present. We thus have an enriched graph consisting of directed and bidirected edges for input to the RnR solver.

\subsection{Density Constraint for Realistic Graphs}
To ensure biological plausibility, we introduce a density constraint that keeps the connectivity density of the recovered graph close to that observed in real functional brain networks. Specifically, the total number of edges in the graph is constrained to be within a realistic range derived from empirical fMRI studies. This prevents solutions that are excessively dense or sparse, when compared to actual brain connectivity.

\begin{figure}[t]
     \includegraphics[width=1\columnwidth]{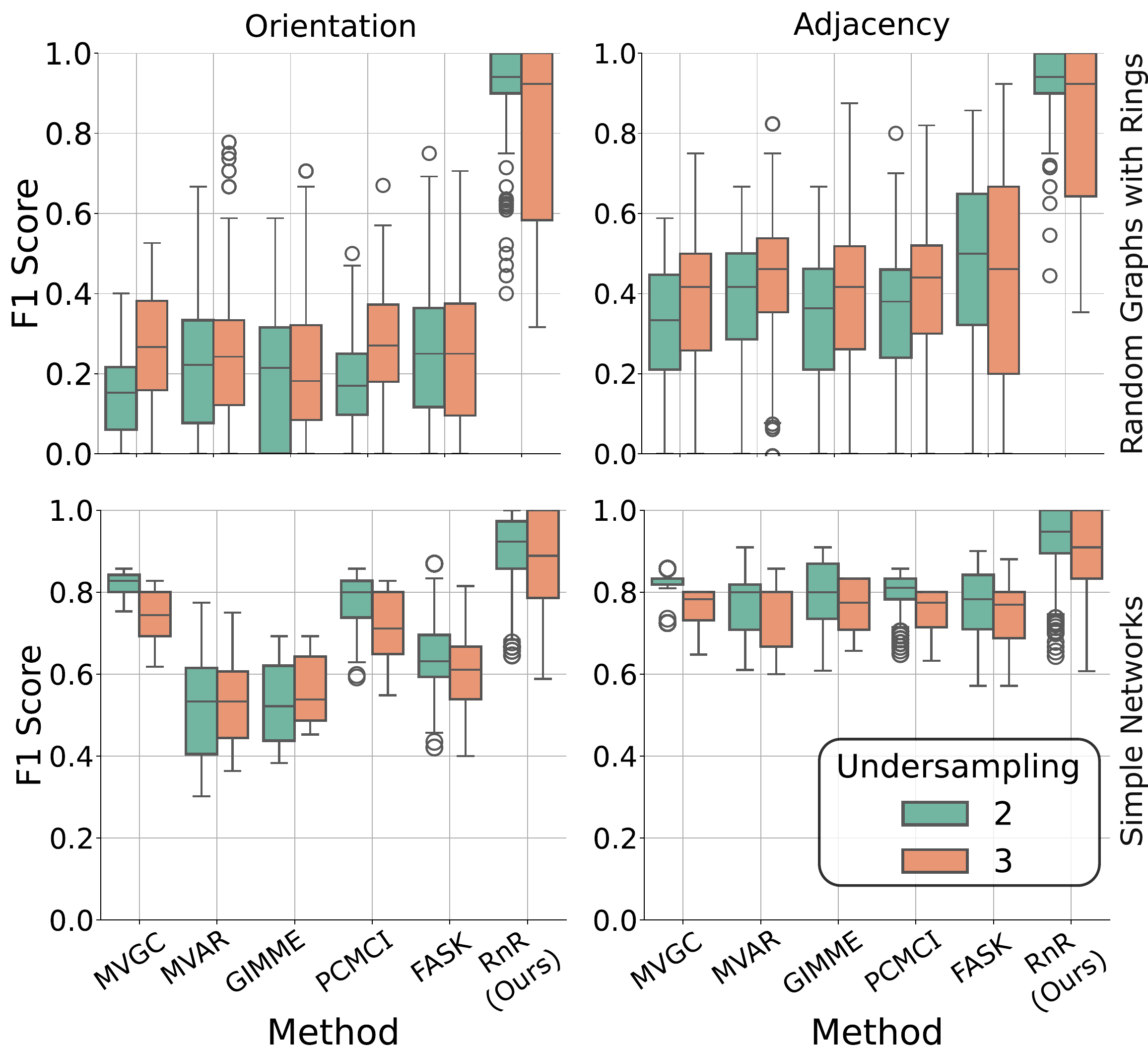}
     \caption{Comparison of Sanchez-Romero's simple data generation approach with larger, more complex VAR-generated graphs. Sanchez-Romero's data lacks real loops and is limited in scope, which may affect the generalizability of causal inference results.}
     \label{fig_var_ringmore}
\end{figure}

\subsection{Prioritized Optimization by Connection Type}
We implemented a multi-stage optimization process that prioritizes specific types of connections to ensure that broad structural characteristics (e.g., density and bidirectional edges) are recovered first, with finer details added in subsequent stages. The optimization first focuses on ensuring that the graph density matches the expected range. Once the density is matched, the algorithm prioritizes recovery of bidirectional connections, which are more likely to represent strong functional interactions, and are the more reliable constraints. In addition, bidirected edges provide important information for inferring the correct undersampling rate. Finally, optimization refines the recovery of directed connections to complete the graph structure. 

\subsection{Adaptive Weighting Schemes}
To handle the inherent uncertainty in fMRI data, we used an adaptive weighting scheme. Edges present in the graph are assigned weights based on connection strength derived from time-series correlations, while absent edges receive maximum weight indicating high confidence in their absence.

\subsection{Practical Justification and Results}
To validate our approach, we applied the modified sRASL to 300 randomly generated graphs undersampled at rates 2, 3, and 4. Figure~\ref{fig_eq} summarizes our findings:

\begin{itemize}
    \item Top Left (A): Solution sets returned while in OptN mode are manageable in size ($\approx 50$), enabling practical manual inspection.
    \item Bottom Right (B): A positive correlation between optimization cost and edge commission error is observed. However, we also find solutions with low errors despite higher costs, highlighting the importance of exploring multiple near-optimal solutions. The two examples (green and red) demonstrate that a top-10 solutions approach based on optimization cost returns the best answer robustly. Thus, we assume that we can identify the best answer, once we have a set of solutions. 
\end{itemize}

\noindent These results justify our choice of OptN mode for optimization, as it provides a richer solution set while maintaining practical usability. 

In conclusion, by leveraging OptN optimization, density constraints, and prioritized optimization strategies, our RnR framework achieves more accurate and realistic recovery of functional brain networks from fMRI data. These modifications address critical limitations in the original method, making it better suited for noisy and complex real-world applications.

\section{Results}

\subsection{Improved Edge-Breaking Experiment}

We replicated and extended the edge-breaking experiment from Abavisani et al. (2023), demonstrating the robustness of the sRASL approach when applied to graphs with intentionally broken edges. In this experiment, we generated a causal ground truth graph \( \mathcal{G}^{1} \) and undersampled it at various rates, simulating noise by randomly deleting an edge. The goal was to test the ability of sRASL to recover the true graph structure. Our results, depicted in Figure~\ref{fig_edge_break}, show that sRASL consistently achieved lower omission and commission errors compared to the original approach, even under high undersampling conditions. This illustrates that sRASL can more effectively restore the true graph structure, making it robust against edge-breaking perturbations.

\subsection{BOLD simulation data: Enhancing Causal Inference with RASL as a Meta-Solver}

We further evaluated the sRASL approach on simulated data from Sanchez-Romero et al. (2019a). Dataset from Sanchez-Romero et al. (2019a), which is widely accepted within the Neuroscience community, was generated with small, simple graphs lacking real loops and using an idiosyncratic data generation method. We demonstrated that Sanchez-Romero’s existing methods do not account for the effects of undersampling, which can bias causal inference.

Our approach applied sRASL as a meta-solver: after running Sanchez-Romero’s original methods, we utilized sRASL on the resulting graph to incorporate the effects of undersampling explicitly. This adjustment led to improved accuracy, as sRASL optimized the causal graph structure to better reflect the underlying dynamics. In essence, sRASL served as an enhancement layer, correcting for undersampling effects ignored by previous methods (Figure~\ref{fig_ruben_all}).

Additionally, we explored the use of PCMCI as an alternative to standard methods like SVAR and Granger Causality (GC). In their 2017 study, Cook, Danks, and Plis demonstrated that SVAR and GC perform well for scenarios with isochronal bidirected edges, which arise due to undersampling, as well as directed edges. However, we observed that PCMCI, initially introduced by Moneta et al. and later significantly improved by Runge, performs better in these scenarios. Therefore, we incorporated PCMCI with RASL, achieving more accurate results by effectively handling undersampling and improving causal inference. Figure~\ref{fig_ruben_all} illustrates these improvements.

\begin{figure*}[ht]
     \includegraphics[width=2\columnwidth]{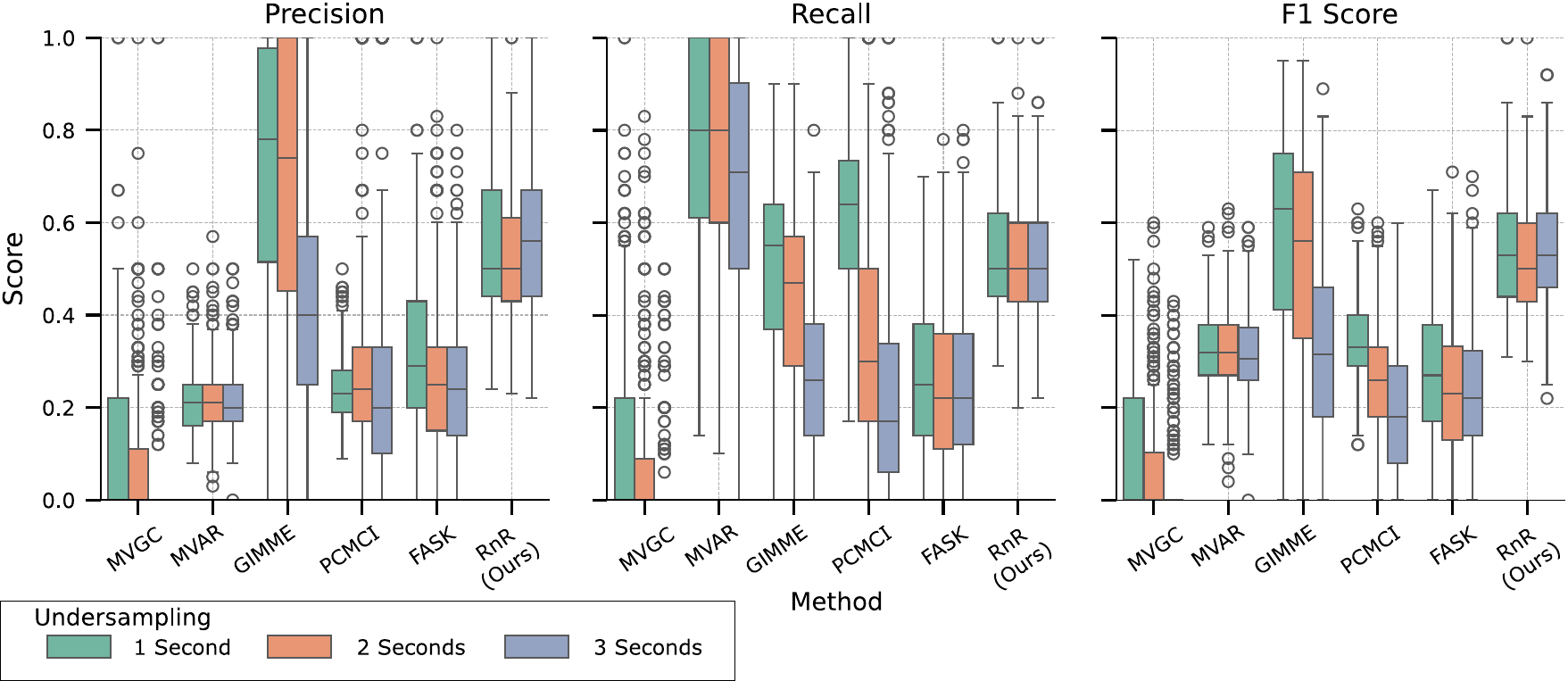}
     \caption{Impact of different undersampling rates (1s, 2s, 3s) on BOLD signal preservation in fMRI data simulated with the balloon model. Minimal error is observed with 1-second undersampling, whereas larger intervals degrade accuracy in all methods that don't account for undersampling effect. Our method RnR accounts for this effect and does not suffer loss from undersampling.}
     \label{fig4}
\end{figure*}
\subsection{Sanchez-Romero's Simulation Data}


\subsection{Analyzing the Limitations of Sanchez-Romero's Data Generation}

Despite its widespread acceptance, Sanchez-Romero's data generation approach is limited in scope and complexity. It employs simple, small graphs without real loops and is generated using a specific, contrived process. To demonstrate the limitations of this dataset for broader causal inference applications, we conducted experiments with larger VAR models on graphs with many variables and more complex structures, showcasing the limitations of Sanchez-Romero's setup (Figure~\ref{fig_var_ringmore}). This underscores the importance of considering more realistic and complex data when testing causal inference methods.


\subsection{Impact of Undersampling on BOLD Signal in fMRI Data}

We also examined the effect of undersampling on time series data generated from a VAR model and then processed through a balloon model to simulate the BOLD response in fMRI data \cite{buxton1998dynamics}. Given the smooth nature of the BOLD signal, undersampling by one-second intervals often leads to minimal information loss, while undersampling by larger intervals (e.g., two or three seconds) introduces significant inaccuracies. Figure~\ref{fig4} illustrates how different undersampling rates impact the preservation of temporal information in BOLD data, underscoring that aggressive undersampling can obscure meaningful connectivity patterns.




\section{Conclusion}

Undersampling is a critical yet often overlooked issue in the analysis of time series data, particularly in the context of fMRI.
In this paper, we addressed the undersampling problem directly by incorporating its effects into the derivation of causal graphs using ASP. Our approach goes beyond traditional methods by explicitly accounting for the temporal disconnect between neural processes and fMRI sampling rates. By doing so, we not only identified the most probable causal graph but also provided an equivalence class of potential graphs, enabling a more nuanced understanding of the underlying causal structures.

Our results, validated on  simulated fMRI data underscore the importance of addressing undersampling in neuroimaging studies. We demonstrated that our ASP-based method outperforms existing techniques, particularly in scenarios where undersampling distorts the true causal relationships between neural groups. Additionally, by comparing our approach to other algorithms, including those studied by~\citet{sanchez2019estimating}, we highlighted that while methods like FASK show potential, they still fall short in fully capturing the complexities introduced by undersampling.

The undersampling issue is not merely a technical challenge but a fundamental barrier to accurate causal inference in neuroscience. Ignoring it risks drawing erroneous conclusions about brain function and connectivity. Our work represents a significant advancement in this area, providing a robust and scalable solution that can improve the accuracy of causal inference in the presence of undersampling.

As the scientific community continues to explore the neural mechanisms underlying cognition and behavior, it is imperative that the undersampling problem is given the attention it deserves. Addressing this issue will be crucial in ensuring that the inferences drawn from fMRI data truly reflect the underlying neural dynamics, leading to more reliable and meaningful insights into brain function.

\section{Future Directions}

One of the challenges of solving the optimization problem with ASP is
the initial graph estimate $\cal{H}$. The estimation errors at the
measurement time-scale may inflate the estimation errors at the causal
timescale. However, simply selecting the estimator that minimizes the
errors in $\cal{H}$, as we have done in this paper, may not be the
optimal strategy. Not all errors in $\cal{H}$ have the same effect on
the quality of estimation and developing methods that consider that
interplay is a promising future direction.

Further optimization of the approaches to enable work with larger graphs may open potential new avenues of where our methods may be applied. Although, our presented approach is highly capable for working with reasonably sized graphs, extending the number of nodes by an order of magnitude could broaden the range of potential applications. This includes futher practical application of the methods in the study of brain function via fMRI as well as other dynamic modalities.

\section{Acknowledgments}
This work was supported by NIH R01MH129047 and in part by NSF 2112455, and NIH 2R01EB006841.
We are grateful to Antti Hyttinen, Matti J\"arvisalo, and Frederick Eberhardt for discussions on clingo.

\bibliographystyle{named}
\bibliography{ijcai25}

\end{document}


\maketitle

\appendix
\section{Appendix}
\label{sec:reference_examples}

\subsection{Brief Introduction on {\tt clingo} and Answer Set Programming (ASP)}
\label{appx:f}
{\tt clingo}~\citep{gebser2011potassco} combines a grounder \texttt{gringo} and a solver \texttt{clasp}. {\tt clingo} is a declarative programming system based on logic programs and their answer sets, used to accelerate solutions of computationally involved combinatorial problems. The grounder converts all parts of a {\tt clingo} program to ``atoms,'' (grounds the statements) and the solver finds ``stable models.'' In ASP, the answer set is a model in which all the atoms are derived from the program and each ``answer'' is a stable model where all the atoms are simultaneously true.

A general {\tt clingo} program includes three main sections, which we show
below using our algorithm as an example:
 \paragraph{Facts:} these are the known elements of the problem. For
  example, the input to Listing~\ref{sRASL} is a graph for which we
  know the edges. A directed edge from node 1 to node 5 is in
  $\mathcal{H}$ translates to \texttt{\textit{hdirected(1,5)}} (line
  1) or if node 1 is part of the SCC number 2, we state this fact in
  {\tt clingo} by \texttt{\textit{scc(1,2)}} (line 2).

\paragraph{Rules:} much like an {\tt if-else} statement, a rule in
  {\tt clingo} consists of a body and a head, formatted as \mbox{\texttt{\textit{head :- body.}}} If all the literals in the body are true, then the head must also be true. Rules can include variables (starting with capital letters), and they are used to derive new facts after grounding. For example:
\begin{equation}
 \texttt{\textit{directed(X, Y, 1) :- edge1(X, Y).}}
\end{equation}
means that for any instantiations of the variables $X$ and $Y$, if we have an edge from $X$ to $Y$, there is a directed path from $X$ to $Y$ of length 1. Before this line, if the model contained the fact \mbox{\texttt{\textit{edge1(2,3)}}}, this line would generate a new fact: \mbox{\texttt{\textit{directed(2,3,1)}}}.

Another type of rule is the ``choice rule'' that describes all the possible ways to choose which atoms are included in the model. For example, in line $5$ of Listing~\ref{sRASL} we used a choice rule to state that the undersampling rate \texttt{\textit{u}} can be anything from 1 to \texttt{\textit{maxu}}. The cardinality constraint:
\begin{equation}
 \texttt{\textit{\{u(1..20)\}.}}
\end{equation}
will generate $2^{20}$ different models (they will not all actually be generated if they conflict with other predicate in each model, or else it would not be possible). In each of these $2^{20}$ models, one subset of all possible atoms generated with this choice rule exists ($\phi$, {\texttt{\textit{\{u(1)\}}}}, {\texttt{\textit{\{u(1), u(2)\}}}}, …). An example of an unconstrained choice rule is line 6 in Listing~\ref{sRASL}, where we want to generate one model for each possible way edges can be present in a graph between two nodes $X$ and $Y$. We can also limit the choice rule. In our problem, only one undersampling rate is present at each solution. We limit the cardinality constraint to have only one member in each model:
\begin{equation}
 \texttt{\textit{1 \{u(1..20)\} 1.}}
\end{equation}
the $1$ on the left is the minimum instantiations of this atom in the model and the $1$ on the right is the maximum. Therefore, we only generate $\binom{20}{1} = 20$ models with this rule, namely one for each undersampling rate. Having several choice rules will multiply the number of generated models by each choice rule.

\textbf{Integrity Constraints:} if choice rules are to generate new models, integrity constraints are there to remove the wrong models from the answers set. More specifically, an integrity constraint is of the form:
\begin{equation}
 \texttt{\textit{:- L0, L1, … .}}
\end{equation}
where literals ${L_{0}, L_{1}, ... .}$ cannot be simultaneously positive. For example, in line $16$ of Listing\ref{sRASL}, we have:
  \begin{equation}\label{len16}
  \begin{aligned}
\texttt{\textit{:- edge1(X, Y), scc(X, K), scc(Y, L), K != L,}}\\
\texttt{\textit{sccsize(L, Z), Z > 1, not dag(K,L). }}
 \end{aligned}
\end{equation}

for cases where the graph consists of several SCCs that are connected using a DAG. If the SCCs are connected by a cyclic directed graph, then the whole graph will become one big Strongly Connected Component. Integrity constraint \ref{len16} states that if there is not a directed edge from a node in SCC K to a node in SCC L as part of the initial DAG, there cannot be such \texttt{\textit{edge1(X, Y)}} from node X to node Y, if node X is in SCC K and node Y is in SCC L.

\subsection{Clingo optimization options}
\label{sec:clingo_options} 
After looking at \texttt{Clingo} options, I found out that we can use the \texttt{optN} option to give us all the optimized solutions. In Section~\ref{Synthetic_Graph_Data} since there are no weights, the optimization cost is very low and there can be many solutions with optimization cost 1. Therefore, if we use normal \texttt{opt} option, it will give us back the first optimum solution it found, whereas there can be many such solutions. 

- when we use \texttt{--opt-mode=enum,4} it will stop searching for solutions as soon as it finds a solution with cost 4 .

- When we use \texttt{--opt-mode=optN,1} it will find all solutions with cost at most 1

- If we use \texttt{--opt-mode=enum,4} coupled with \texttt{-n 100} it will give us 100 answers with cost at most 4. If we use \texttt{-n 0} which is CAPSIZE zero it will give us all the answers that have cost less than 4.

-If we use \texttt{--opt-mode=optN,3 -n 4} it will look for solutions with cost less than 3, but it doesn't stop there.It continues to find the optimal answer, lets say it has a cost of 2, then it will enumerate and give us 4 answers with the optimal cost 2.

- If we run with \texttt{--opt-mode=opt -n 0} even though the CAPSIZE is zero,  it will only give us one optimal answers.

- If we run Clingo optimization on an exact undersampled graph that has no error, and if we use \texttt{--opt-mode=optN} it will find solutions with cost zero, which are the correct G1s with regard to that h. Therefore gives us the complete equivalent class.

\subsection{Clingo configuration} 
\begin{figure*}[ht]
     \includegraphics [width = 2\columnwidth]{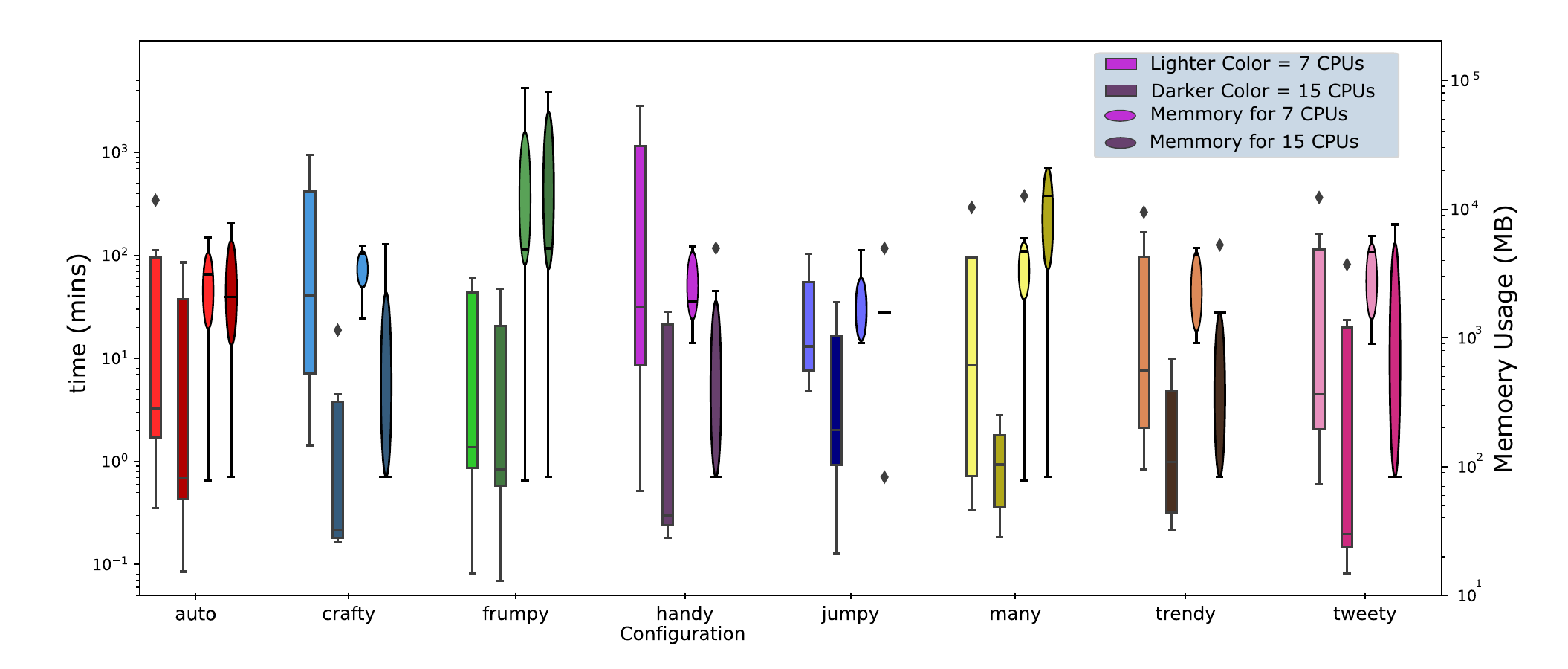}
     \caption{Time to solve and memory usage of different Clingo configurations over two settings of using 7 or 15 CPUs. Rectangular boxes show time behavior and oval boxes show memory behavior. Darker color in each category is for 15 CPUs and lighter color is for 7 CPUs.}
     \label{fig6}
\end{figure*}

\begin{figure}
     \includegraphics [width = 1\columnwidth]{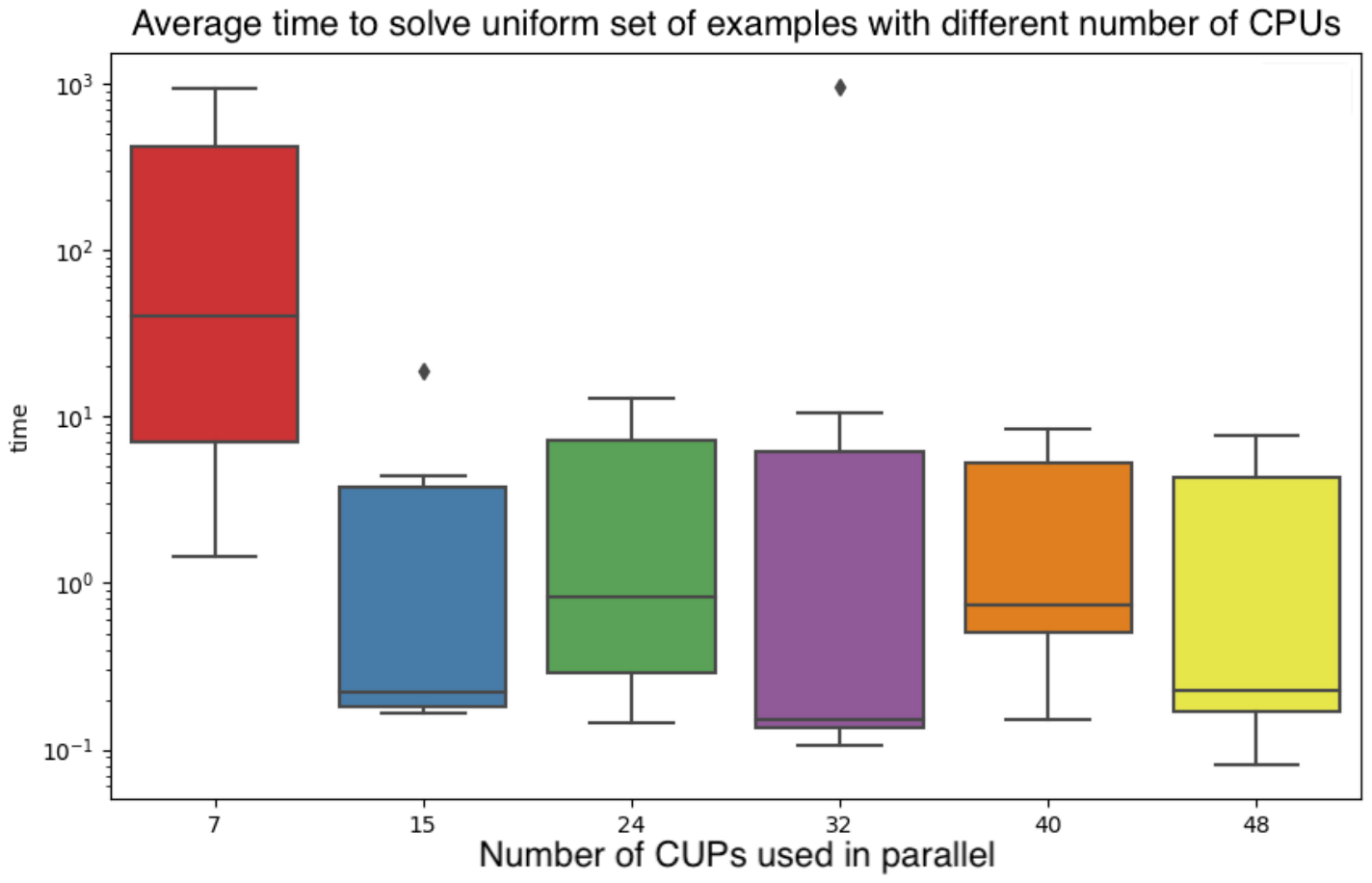}
     \caption{Time to solve same examples with different number of CPUs working in parallel.}
     \label{fig7}
\end{figure}

Since there are many options to run Clingo with and each have their own appropriate problem to solve, here we wanted to find the best fit for our problem. To this end, we ran the same set of unified test sets on different configurations and measured their statistics on time to solve and memory usage. We also experimented with different number of CPUs to find the optimal setting for our problem. Figure~\ref{fig6} summarizes these findings. Based on out experiments, using Crafty with 15 CPUs resulted in the fastest and most memory efficient runs. However, these configurations are problem dependent and other problems might behave better with different configurations. The reason we did not go further than 15 CPUs is that out experiments show adding more CPUs in parallel does not speed up solving in general. We show these results in Figure~\ref{fig7}

\subsection{VAR results 3 way to optimize}
In VAR simulation, we have 3 ways to report the error of our solution:

\begin{enumerate}
    \item Error between the optimal solution Clingo has found, undersampled at rate $u$, and ground truth undersampled at $u$. We call it $G^{u}_{\text{opt}} \text{ Vs. } GT_{u}$.

    \item Error between the optimal solution Clingo has found, undersampled at rate $u$, and the estimate resulting after SVAR. We call it $G^{u}_{\text{opt}} \text{ Vs. } G_{\text{estimate}}(=\mathcal{H})$.

    \item Error between the optimal solution Clingo has found, undersampled at rate 1, and ground truth at rate 1. We call it $G^{1}_{\text{opt}} \text{ Vs. } GT$.
\end{enumerate}

When I run Clingo with optN, I will have a set of optimal solutions that I have to loop through and find the minimum error. But minimize with respect to which one of these 3?

In order to answer this question, I looped through all the answers 3 times and found the optimal answer with respect to either one of these 3. For that particular answer, I computed all these 3 errors. This figure summarizes this experiment.

Basically, I want to find out which of these 3 errors I should use to find the optimal answer with respect to. Figure~\ref{fig8} shows these results. 
\begin{figure}
     \includegraphics [width = 1\columnwidth]{figs/VAR_3_ways_initial.png}
     \caption{VAR simulation results on 8-node random graphs and all the ways to report errors.}
     \label{fig8}
\end{figure}

The most informative columns is column3 out of 9 columns from right. It shows how actuate we were able to estimate a solution close to ground truth without knowing the ground truth in looping through all the optimal solutions.

\bibliographystyle{named}
\bibliography{supp_bib}